\documentclass{article}
\usepackage{spconf,amsmath,graphicx}

\usepackage{microtype}
\usepackage{multirow}
\usepackage{float}
\usepackage{tabulary}
\usepackage{xcolor}
\usepackage{hyperref}

\newcommand{\citep}[1]{\cite{#1}}
\newcommand{\citet}[1]{\cite{#1}}
\newcommand{\citealp}[1]{\cite{#1}}

\title{KG-ECO: Knowledge Graph Enhanced Entity Correction \\ for Query Rewriting}

\name{Jinglun Cai*\thanks{*This work was completed while the first author was an intern at Amazon.}, Mingda Li, Ziyan Jiang, Eunah Cho, Zheng Chen, Yang Liu, Xing Fan, Chenlei Guo}
\address{Amazon.com, Inc. USA}
\begin{document}
%
\maketitle

\begin{abstract}

Query Rewriting (QR) plays a critical role in large-scale dialogue systems for reducing frictions. When there is an entity error, it imposes extra challenges for a dialogue system to produce satisfactory responses. In this work, we propose \textbf{KG-ECO}: \textbf{K}nowledge \textbf{G}raph enhanced \textbf{E}ntity \textbf{CO}rrection for query rewriting, an entity correction system with corrupt entity span detection and entity retrieval/re-ranking functionalities.To boost the model performance, we incorporate Knowledge Graph (KG) to provide entity structural information (neighboring entities encoded by graph neural networks) and textual information (KG entity descriptions encoded by RoBERTa). Experimental results show that our approach yields a clear performance gain over two baselines:  utterance level QR  and  entity correction  without utilizing KG information. The proposed system is particularly effective for few-shot learning cases where target entities are rarely seen in training or there is a KG relation between the target entity and other contextual entities in the query. 
\end{abstract}


\section{Introduction}

Large-scale conversational AI based dialogue systems like Alexa, Siri, and Google Assistant, serve millions of users on a daily basis. Inevitably, some user queries result in frictions or errors. Such frictions may originate from either the dialogue system itself, or user ambiguity.
Query Rewriting (QR) aims to automatically rephrase a user query into another form. For instance, the upstream module in a dialogue system may produce a query with a wrong song name: ``play bad boy dance by lady gaga''. The QR system should rewrite the utterance into ``play bad romance by lady gaga''.



In query rewriting,  corrupt entity correction\footnote{We focus on entity correction for textual inputs. Various related terms are used in literature, including entity retrieval / resolution / understanding with noisy input.} can be challenging 
when we have no knowledge about the entities and limited context information. To address this, we utilize an external knowledge graph, namely Wikidata\footnote{https://www.wikidata.org}, to facilitate the entity correction task. Typical functionalities for a conversational AI include playing music, playing videos, reading books and weather forecast. We found Wikidata can cover most common entities in user queries such as celebrity names, artwork names, media names and locations.





\begin{figure*}[ht]
\centering
\includegraphics[width=17cm, height=6.5cm]{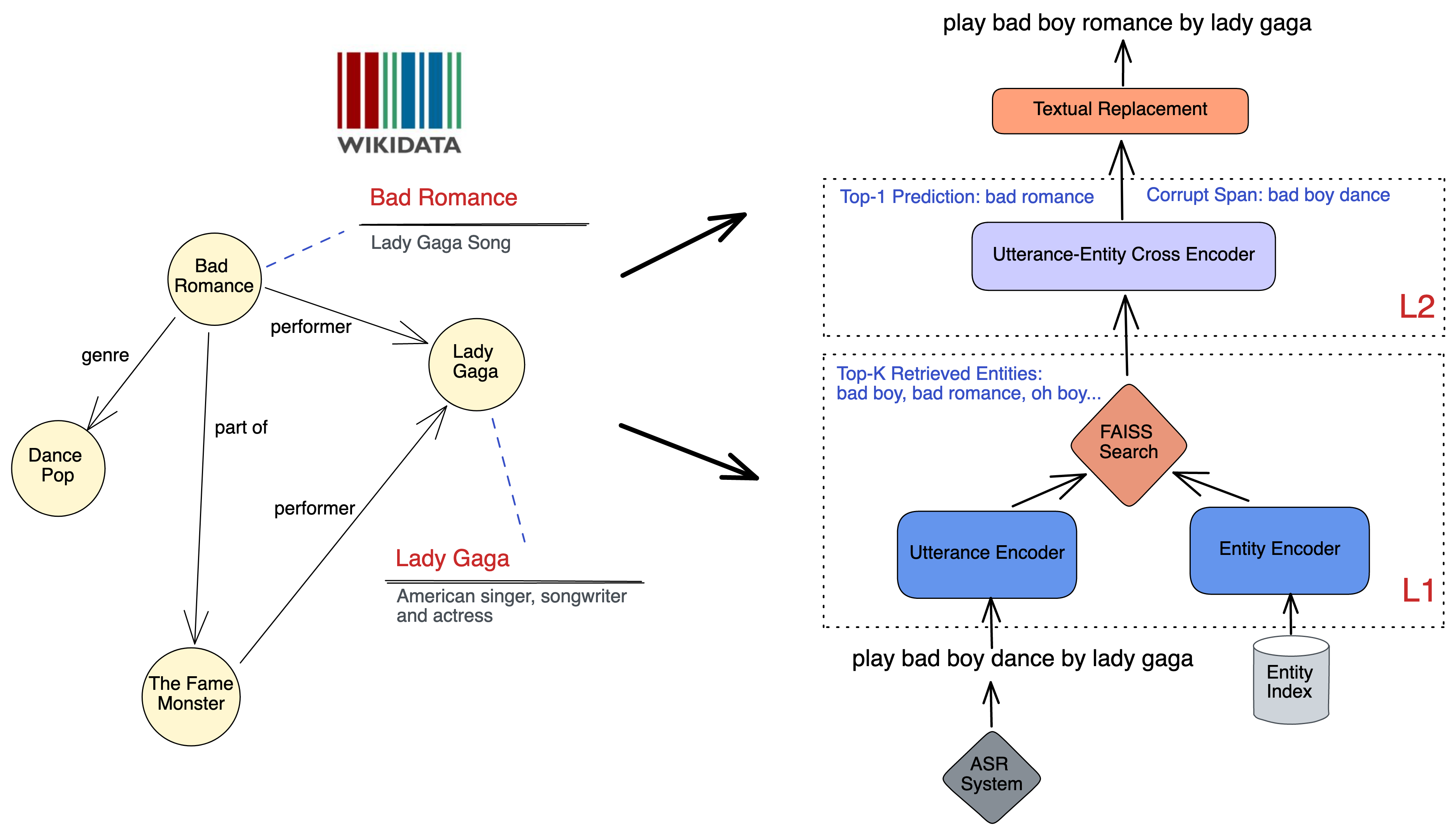}
\caption{KG-ECO System Overview. The input query ``play bad boy dance by lady gaga'' is the output from upstream ASR system which contains one corrupt entity ``bad boy dance''. Our KG-ECO system, which has a two-stage architecture, predicts the correct entity ``bad romance'' together with the span of corrupt entity. The final rewrite is produced through a textual replacement. KG information is utilized by our system.}
\label{fig: System Architecture}
\end{figure*}


In this work, we introduce a two-layer approach: retrieval (L1), and re-ranking + span detection (L2). 
In the second layer we jointly re-rank the entity candidates and detect whether and where a corrupt/erroneous entity span exists in the query. The span detection task is critical for ensuring the flexibility of the QR system, as we do not assume any upstream Natural Language Understanding (NLU) or Named-entity recognition (NER) module providing entity candidates. In both L1 and L2, we incorporate the KG text description of entities, 
 and leverage the Graph Attention Network (GAT) \citep{Velickivic2018} to encode the KG structural information.
Evaluations are conducted on a friction set and a clean set, representing different scenarios of real-world applications. 
The effectiveness of our system is demonstrated by a clear performance gain over two baselines: an utterance level QR system (gUFS-QR) \citep{Fan2021} and an entity correction QR system without KG information.

\section{Related work}
Existing efforts treat QR as text generation problem\citep{su2019improving} or retrieval problem\citep{Fan2021,Chen2020,Cho2021,naresh2022pentatron}. Entities have been shown to be a strong indicator of text semantics and be critical to QR task\citep{naresh2022pentatron}. 
With particular emphasis on entities, we can also easily leverage Knowledge Graphs (KG), which  provides rich information about entities. Our work is the first effort to utilize KG information in the QR task.



Our work is also closely related to two entity-level tasks: entity correction and entity linking. Entity correction \citet{Raghuvanshi2019, WangH2020, Muralidharan2021, WangH2021} aims to tackle errors occurring in Automatic Speech Recognition (ASR) systems. 
These studies adopt an ``entity-to-entity'' approach for entity correction, however, we take the query as context, and perform contextualized entity correction. Meanwhile, we do not assume we know the location of the corrupt entity if it exists. Thus we perform entity span detection jointly with entity re-ranking. Entity linking \citep{wu2019zero,li2020efficient} is another similar task which aims to link mentioned entities with their corresponding entities in a knowledge base. Our work is more challenging because the input utterance is noisy with incorrect entities.

\section{KG-ECO}

Our Knowledge Graph enhanced Entity COrrection system (KG-ECO) consists of two layers: retrieval (L1) and re-ranking + span detection (L2), as illustrated in Figure \ref{fig: System Architecture}. For each utterance, we efficiently retrieve top-K relevant entities\footnote{In this work, we retrieve and re-rank entities in surface form (the textual name of an entity), as the resolved entity labels are NOT available in query rewriting data. In case of \textit{polysemy} (one surface form corresponds to multiple KG entities), we incorporate multiple entities into the input.}
encoded by the L1 model (a bi-encoder model) from the entity index. Then, the L2 model (a cross encoder model) re-ranks the top-K retrieved candidates, and detects a corrupt span (possibly empty) from the input utterance.
Finally, under optimized triggering conditions, the top ranked entity will be used for rewriting via a textual replacement.


\subsection{L1 Retrieval}
For the L1 retrieval model, we adopt a bi-encoder architecture, which comprises an utterance encoder and an entity encoder. 
The utterance encoder $\textrm{E}_{\textrm{utt}}$ is a RoBERTa based model, accepting the source utterance as the input. The entity encoder $\textrm{E}_{\textrm{ent}}$ consists of a RoBERTa based ``Entity\_Description'' encoder, and a GAT encoder, which consumes one-hop subgraphs as its input, described in details later. The relevance between an utterance $p$ and an entity $q$ is defined by their dot product similarity: 

\vspace{-3mm} 
\begin{equation}
\textrm{sim}(p, q) = \textrm{E}_{\textrm{utt}}(p)^T \cdot \textrm{E}_{\textrm{ent}}(q)
\end{equation}
\vspace{-3mm} 

\textbf{Training.} We utilize Negative Log-Likelihood (NLL) as the loss function in training such that relevant utterance and entity pairs will have higher dot product similarity than negative pairs. 
Positive entities (ground truth) were specified when data sets were constructed. 
To obtain negative entities, we used two approaches: in-batch negatives and hard negatives. In-batch negatives are the other entities in the same training batch that are not positive. In this way, we efficiently utilize the computation of entity embeddings for an entire batch. 
For hard negatives, we follow \citet{Gillick2019} and \citet{Karpukhin2020} to use hard negative entities in retrieval training.
For instance, the positive entity ``carson city'' has its hard negative ``corbin city''. This helps the retrieval layer to distinguish an entity from its highly similar competitors in the entity index. 

\textbf{Inference.} After training, we first produce entity embeddings using the entity encoder, and build an index via FAISS \citep{Johnson2017}, a scalable similarity search framework for dense vectors. In inference, given an utterance, we obtain its embedding and conduct FAISS search from the index to retrieve the top-K most similar entities in terms of dot product. These top-K entities are candidates for the next stage. 

\subsection{L2 Re-ranking + Span Detection}

The L2 architecture consists of a RoBERTa based cross encoder and a GAT encoder. The cross encoder consumes both the utterance and the entity and its description as input. The L2 layer is a joint learning model with two learning tasks: re-ranking and span detection.

For re-ranking, given a pair of utterance and entity, we concatenate the output vector of CLS token of RoBERTa and the pooling output vector of GAT, and pass them to an MLP layer to produce the relevance score of the pair. For corrupt entity span detection, we predict the span's start and end positions at the token level, following  similar approaches such as in \citet{Devlin2018} and \citet{WangZ2021}. Specifically, assume $W_S$ and $W_E$ are the start and the end vector respectively,  and $T_i\in R^H$ is the final hidden vector for the $i^{th}$ input token, then the score of a candidate span from position $i$ to position $j$ is computed as: 

\vspace{-3mm} 
\begin{equation}
s_{ij}=W_S\cdot T_i+W_E\cdot T_j 
\end{equation}

\vspace{-1mm} 

In addition, we introduce a special case: a null span, which means that no corrupt entity exists. 
This happens when $i = j = 0$, i.e., the start and end tokens are both the CLS token of RoBERTa.
We select a threshold $\theta$ for null prediction, by balancing the precision and false trigger rate.



\subsection{KG Enhanced Component: Entity Description}
KG provides short textual descriptions of entities. We use these to augment the textual input: we concatenate an entity and its descriptions, separated by special token [des]. For example, entity ``bad romance'' is polysemic, corresponding to two KG entities, a song and a film. We concatenate both descriptions to obtain the input ``bad romance [des] song [des] 2011 film''. The description ``song'', will help the system to learn the relevance between utterance ``play bad boy dance by lady gaga'' and entity ``bad romance''. 

\subsection{KG Enhanced Component: GAT}


We incorporate Graph Attention Network (GAT) \citep{Velickivic2018}, a state-of-the-art Graph Neural Network architecture, as a component in both L1 and L2 modules. Assume $(h_0, ..., h_n)$ is a sequence of graph node embeddings. A GAT layer outputs a transformed sequence $(h_0', ..., h_n')$. Assume $N_i$ is the neighborhood of node $i$. 

\vspace{-5mm} 
\begin{equation}
 \alpha_{ij}  = \frac{\exp(\textrm{LeakyReLU}(a^T[Wh_i || Wh_j]))}{ \Sigma_{k\in N_i}\exp(\textrm{LeakyReLU}(a^T[Wh_i || Wh_k]))}
\end{equation}

\vspace{-2mm} 

\begin{equation}
h_i' = \sigma (\Sigma_{j\in N_i}\alpha_{ij} W h_j )
\end{equation}

\vspace{-1mm} 

\noindent where $||$ represents concatenation, and $\sigma$ is a non-linear activation.

In this work, we use the one-hop subgraph of the target entity as the input to GAT. 
To be specific, the input KG embeddings, including both node and relation embeddings, are pre-trained with the link prediction objective. These input embeddings are fixed during training. 


The vanilla GAT design does not support relation-type learning, and only transforms node embeddings. Inspired by \citet{Vashishth2020}, we introduce a non-parametric composition of node and relation embeddings. Suppose $(i, r, j)$ is a KG triple, where $i,j$ are nodes and $r$ is the relation. We apply a non-parametric function $\phi$ to merge the relation and node embeddings $(h_r, h_j)$ while updating $h_i$. We experimented with both subtraction and product for function $\phi$, and finally selected subtraction.

\begin{equation}
h_i' = \sigma (\Sigma_{(r,j)\in N_i}\alpha_{ij} W_{\mathrm{node}}\phi (h_r, h_j) )
\end{equation}

Meanwhile, the relation embedding $h_r$ is also updated via a dedicated MLP layer: $h_r' = W_{\mathrm{rel}} h_r$.




\section{Experiments}


\subsection{Data}
\textbf{Knowledge Graph Datasets.}
Our entity correction system is enhanced by Wikidata, one of the largest publicly available knowledge bases. Specifically, we use the following two data artifacts: \textit{(I) KG Structure/Description.} Kensho Derived Wikimedia Dataset\footnote{https://www.kaggle.com/kenshoresearch/kensho-derived-wikimedia-data} is a pre-processed English Wikidata set. It contains 44M entities  and 141M relation triples. This data set serves as a rich source of external knowledge of graph structure and textual description. \textit{(II) Retrieval Entity Index.} 
To increase inference efficiency,  we use a more condensed entity set to build the retrieval entity index. We start with a derived Wikidata version provided by Spacy Entity Linker\footnote{https://github.com/egerber/spaCy-entity-linker}, and further remove entities with digits/punctuation/non-ascii symbols.
The resulting entity index is 3M in surface form (textual name). 


\textbf{Rephrase Datasets.} 
Rephrase pairs are gathered from the production traffic of three months in a large-scale dialogue system. We remove all private and sensitive information in the  utterances, and ensure our data sets are de-identified. Rephrase pairs are consecutive user queries where the first turn contains friction and the second turn is successful based on a friction detection model, following the methods described in \citealp{Cho2021}. Corrupt and target entities (if they exist) are further extracted using the NLU hypotheses of the rephrase data. 




\textbf{Training Datasets.} 
For L1 Retrieval training, we use a subset (2.4M utterances) of the entire training dataset, by requiring that the corrupt and target entity pair exists and the target entity appears in the entity index. This way we ensure that the model is trained to retrieve most relevant entities from the index. For L2 Re-ranking + Span detection training, we use all the training data (8.5M utterances), so that no-entity-corruption cases can support null span learning.



\begin{table*}[!ht]
\centering

    \begin{tabular}{c|cccc|cccc}
    \hline
    \multirow{2}{*}{KG Component} & \multicolumn{4}{c|}{Entity Precision (E-P)}                                                                                           & \multicolumn{4}{c}{NLU Precision (NLU-P)}                                                \\
      & Overall & Zero-shot & \begin{tabular}[c]{@{}c@{}}Few-shot\end{tabular} & \begin{tabular}[c]{@{}c@{}}KG-relation\end{tabular} & Overall & Zero-shot & \begin{tabular}[c]{@{}c@{}}Few-shot\end{tabular} & \begin{tabular}[c]{@{}c@{}}KG- relation\end{tabular} \\ \hline
    None                                     & 38.7    & 5.6    & 23.0                                                  & 36.8                                                   & 30.6    & 4.4    & 18.8                                                & 27.9                                                   \\
    Description                            & 41.4    & 11.6   & 28.0                                                  & 39.5                                                   & 33.0      & 8.8    & 23.6                                                & 30.4                                                   \\
    GAT + Description                      & \textbf{43.9}   & \textbf{12.3}   & \textbf{31.3}                                                 & \textbf{43.2}                                              & \textbf{34.9}    & \textbf{9.5}   & \textbf{26.4}      & \textbf{33.1}                                                    \\ \hline
    \end{tabular}
    
\caption{Ablation study of KG-ECO on friction subsets. No trigger threshold applied.}
\label{table: Performance of entity correction system on friction subsets}
\end{table*}



\textbf{Test Datasets.} To reflect different scenarios of real-world applications, our test sets include a friction set (107K utterances) and a clean set (3K utterances). In the friction set, each source utterance contains exactly one corrupt entity, and the target entity exists in the entity index.  Three subsets of the friction set are considered: zero-shot, few-shot and KG relation. \textit{zero-shot} set contains data where target entities do not appear in the training set; \textit{few-shot} set contains data with target entities appearing $1 \sim 10$ times; 
in \textit{KG relation} set the target entity has a KG relation with some context entity in the source utterance. The clean set contains utterances that do not need to be rephrased,
and serves as a safety measure: it calibrates how likely a model falsely rewrites a clean input that does not need to be corrected.   


\subsection{Evaluation Metrics and Baseline}
We use the following evaluation metrics\footnote{Trigger rate applies to both friction and clean sets, while the other three metrics only apply to the friction set.}: \textbf{Entity Precision (E-P)}: The fraction of correct rank 1 entity predictions over all the triggered predictions. \textbf{NLU Precision (NLU-P)}: The fraction of correct NLU hypothesis predictions (in terms of exact match) over all triggered predictions. \textbf{Trigger Rate (TR)}: The ratio between rewrite-triggered test samples and all the test samples. \textbf{Correct Trigger Rate (CTR)}: The fraction of correctly triggered NLU hypothesis predictions over all the test samples; i.e., Correct trigger rate = Trigger rate $\times$ NLU precision.
To generate the NLU hypothesis, we first obtain the NLU hypothesis for the source utterance, and replace its corrupt span with the top 1 ranked entity. For the example in Figure \ref{fig: System Architecture}, the system replaces ``bad boy dance'' with ``bad romance'' in the NLU hypothesis ``Music $|$ PlayMusicIntent $|$ ArtistName: lady gaga$|$ SongName: bad boy dance''.



gUFS-QR \citep{Fan2021}, a state-of-the-art QR system, serves as a baseline.
It is a two-layer system, retrieving and re-ranks utterances and NLU hypotheses. 
Since we do not assume the entities are tagged in utterances, ``entity-to-entity'' approaches \citet{Raghuvanshi2019, WangH2020, Muralidharan2021, WangH2021} are not appropriate baselines. Besides gUFS-QR, we also consider our two-layer entity correction system without the KG components, which is a similar design as popular entity linking system\citet{wu2019zero}, as a baseline.

\interfootnotelinepenalty=10000
\subsection{Results}

We first present evaluation results without the trigger threshold, i.e., the rewrite is always triggered. Results of our KG-ECO system on each friction subset are shown in Table~\ref{table: Performance of entity correction system on friction subsets}. We observe that KG is particularly beneficial for zero-shot and few-shot learning. 
This is expected since the KG serves as an external source of information. Moreover, we can see GAT is more effective on the KG-relation set than on the overall friction set.



\begin{table}[!ht]
\centering

    \begin{tabular}{c|cc}
    \hline
    System                   & E-P  & NLU-P \\ \hline
    gUFS-QR                  & 33.4 & 26.7  \\
    Entity Correction w/o KG & 38.7 & 30.6  \\
    KG-ECO                   & \textbf{43.9}     & \textbf{34.9}  \\ \hline
    \end{tabular}

\caption{Performance on friction set. No trigger threshold applied.}
\label{table: Performance on friction set}
\end{table}









\begin{table}[!ht]
\centering
    \begin{tabular}{c|cc|c}
    \hline
    \multirow{2}{*}{System}  & \multicolumn{2}{c|}{Friction Set} & Clean Set \\
           & TR $\uparrow$    & CTR   $\uparrow$   & TR $\downarrow$       \\ \hline
    gUFS-QR                  & 14.9            & 9.7             & 2.4       \\
    Entity Correction w/o KG & 41.8            & 15.4            & \textbf{2.3}       \\
    KG-ECO                   & \textbf{46.2}         & \textbf{18.8}          & \textbf{2.3}        \\ \hline
    \end{tabular}
\caption{Triggered performance on clean and friction sets. 
}
\label{table: Triggered results on test sets}
\end{table}

Table~\ref{table: Performance on friction set} shows the overall performance of our KG-ECO system, compared to gUFS-QR baseline\footnote{By design, gUFS-QR system's predictions are at utterance/NLU hypothesis level. To evaluate entity precision for gUFS-QR, we check if the target entity appears in the slot values of top 1 NLU hypothesis.} and the entity correction system without KG enhanced components. We can see our method achieves the best performance in both entity precision and NLU precision. This shows that leveraging KG benefits the entity correction task substantially. 
Notice that even though the NLU hypothesis prediction is based on a straightforward replacement method, we can already obtain a decent absolute improvement of 8.2\%.

In Table \ref{table: Triggered results on test sets}, we present the results with tuned trigger conditions. We search the null threshold for span detection in $(3, 4, 5, 6, 7)$,  keeping only the results that have smaller (false) trigger rates on the clean set than gUFS-QR. Finally we select the threshold maximizing the correct trigger rate on the friction set.
Our entity correction system achieves significantly higher trigger rates, while gUFS-QR appears to be overly conservative with a low trigger rate. Overall, it is remarkable that the KG-ECO system outperforms gUFS-QR baseline by 9.1\% in terms of the correct trigger rate.

\section{Conclusions}

In this work, we present a novel entity correction approach for Query Rewriting, powered by Knowledge Graph.  The proposed KG-ECO system significantly outperforms two baselines: an utterance level QR system and the entity correction system without KG information, on different datasets,  friction samples and clean samples. In particular, the system is exceptionally effective for few-shot learning, which demonstrates its potential for generalization. 

\bibliographystyle{IEEEbib}
\bibliography{strings,refs}

\begin{thebibliography}{10}

\bibitem{Velickivic2018}
Petar Veli{\v c}kovi{\'c}, Arantxa Casanova, Pietro Li{\`o}, Guillem Cucurull,
  Adriana Romero, and Yoshua Bengio,
\newblock ``Graph attention networks,''
\newblock {\em ICLR}, 2018.

\bibitem{Fan2021}
Xing Fan, Eunah Cho, Xiaojiang Huang, and Chenlei Guo,
\newblock ``Search based self-learning query rewrite system in conversational
  ai,''
\newblock {\em In 2nd International Workshop on Data-Efficient Machine Learning
  (DeMaL)}, 2021.

\bibitem{su2019improving}
Hui Su, Xiaoyu Shen, Rongzhi Zhang, Fei Sun, Pengwei Hu, Cheng Niu, and Jie
  Zhou,
\newblock ``Improving multi-turn dialogue modelling with utterance rewriter,''
\newblock {\em arXiv preprint arXiv:1906.07004}, 2019.

\bibitem{Chen2020}
Zheng Chen, Xing Fan, Yuan Ling, Lambert Mathias, and Chenlei Guo,
\newblock ``Pre-training for query rewriting in a spoken language understanding
  system,''
\newblock {\em ICASSP}, 2020.

\bibitem{Cho2021}
Eunah Cho, Ziyan Jiang, Jie Hao, Zheng Chen, Saurabh Gupta, Xing Fan, and
  Chenlei Guo,
\newblock ``Personalized search-based query rewrite system for conversational
  {AI},''
\newblock in {\em Proceedings of the 3rd Workshop on Natural Language
  Processing for Conversational AI}. Nov. 2021, pp. 179--188, Association for
  Computational Linguistics.

\bibitem{naresh2022pentatron}
Niranjan~Uma Naresh, Ziyan Jiang, Sungjin Lee, Jie Hao, Xing Fan, Chenlei Guo,
  et~al.,
\newblock ``Pentatron: Personalized context-aware transformer for
  retrieval-based conversational understanding,''
\newblock {\em arXiv preprint arXiv:2210.12308}, 2022.

\bibitem{Raghuvanshi2019}
Arushi Raghuvanshi, Vijay Ramakrishnan, Varsha Embar, Lucien Carroll, and
  Karthik Raghunathan,
\newblock ``Entity resolution for noisy {ASR} transcripts,''
\newblock in {\em EMNLP-IJCNLP}, Hong Kong, China, Nov. 2019, pp. 61--66,
  Association for Computational Linguistics.

\bibitem{WangH2020}
Haoyu Wang, Shuyan Dong, Yue Liu, James Logan, Ashish~Kumar Agrawal, and Yang
  Liu,
\newblock ``Asr error correction with augmented transformer for entity
  retrieval,''
\newblock {\em INTERSPEECH}, 2020.

\bibitem{Muralidharan2021}
Deepak Muralidharan, Joel Ruben~Antony Moniz, Sida Gao, Xiao Yang, Justine Kao,
  Stephen Pulman, Atish Kothari, Ray Shen, Yinying Pan, Vivek Kaul, Mubarak
  Seyed~Ibrahim, Gang Xiang, Nan Dun, Yidan Zhou, Andy O, Yuan Zhang, Pooja
  Chitkara, Xuan Wang, Alkesh Patel, Kushal Tayal, Roger Zheng, Peter Grasch,
  Jason~D Williams, and Lin Li,
\newblock ``Noise robust named entity understanding for voice assistants,''
\newblock in {\em NAACL: Industry Papers}, Online, June 2021, pp. 196--204,
  Association for Computational Linguistics.

\bibitem{WangH2021}
Haoyu Wang, John Chen, Majid Laali, Kevin Durda, Jeff King, William Campbell,
  and Yang Liu,
\newblock ``Leveraging asr n-best in deep entity retrieval,''
\newblock {\em INTERSPEECH}, 2021.

\bibitem{wu2019zero}
Ledell Wu, Fabio Petroni, Martin Josifoski, Sebastian Riedel, and Luke
  Zettlemoyer,
\newblock ``Zero-shot entity linking with dense entity retrieval,''
\newblock in {\em EMNLP}, 2020.

\bibitem{li2020efficient}
Belinda~Z. Li, Sewon Min, Srinivasan Iyer, Yashar Mehdad, and Wen-tau Yih,
\newblock ``Efficient one-pass end-to-end entity linking for questions,''
\newblock in {\em EMNLP}, 2020.

\bibitem{Gillick2019}
Daniel Gillick, Sayali Kulkarni, Larry Lansing, Alessandro Presta, Jason
  Baldridge, Eugene Ie, and Diego Garcia-Olano,
\newblock ``Learning dense representations for entity retrieval,''
\newblock in {\em CoNLL}, Hong Kong, China, Nov. 2019, pp. 528--537,
  Association for Computational Linguistics.

\bibitem{Karpukhin2020}
Vladimir Karpukhin, Barlas Oguz, Sewon Min, Patrick Lewis, Ledell Wu, Sergey
  Edunov, Danqi Chen, and Wen-tau Yih,
\newblock ``Dense passage retrieval for open-domain question answering,''
\newblock in {\em EMNLP}, Online, Nov. 2020, pp. 6769--6781, Association for
  Computational Linguistics.

\bibitem{Johnson2017}
Jeff Johnson, Matthijs Douze, and Hervé Jégou,
\newblock ``Billion-scale similarity search with gpus,''
\newblock {\em IEEE Transactions on Big Data}, vol. 7, no. 3, pp. 535--547,
  2021.

\bibitem{Devlin2018}
Jacob Devlin, Ming-Wei Chang, Kenton Lee, and Kristina Toutanova,
\newblock ``{BERT}: Pre-training of deep bidirectional transformers for
  language understanding,''
\newblock in {\em NAACL}, Minneapolis, Minnesota, June 2019, pp. 4171--4186,
  Association for Computational Linguistics.

\bibitem{WangZ2021}
Zhuoyi Wang, Saurabh Gupta, Jie Hao, Xing Fan, Dingcheng Li, Alexander~Hanbo
  Li, and Chenlei Guo,
\newblock ``Contextual rephrase detection for reducing friction in dialogue
  systems,''
\newblock in {\em EMNLP}, Online and Punta Cana, Dominican Republic, Nov. 2021,
  pp. 1899--1905, Association for Computational Linguistics.

\bibitem{Vashishth2020}
Shikhar Vashishth, Soumya Sanyal, Vikram Nitin, and Partha Talukdar,
\newblock ``Composition-based multi-relational graph convolutional networks,''
\newblock in {\em ICLR}, 2020.

\end{thebibliography}

\end{document}